\documentclass[preprint,12pt]{elsarticle}

\usepackage{graphicx}%
\usepackage{multirow}%
\usepackage{amsmath,amssymb,amsfonts}%
\usepackage[table]{xcolor}
\usepackage{tabularx}
\usepackage[ruled,vlined]{algorithm2e}
\usepackage{longtable} 
\usepackage{booktabs} 
\usepackage{booktabs}
\usepackage{makecell}
\usepackage{array}
\PassOptionsToPackage{colorlinks=true, linkcolor=blue}{hyperref}
\usepackage{hyperref}
\usepackage{graphicx} 
\usepackage{algorithmic}
\usepackage{amsmath}

\begin{document}
\let\WriteBookmarks\relax
\def\floatpagepagefraction{1}
\def\textpagefraction{.001}

\begin{frontmatter}



\title {Graph Neural Network-based Spectral Filtering Mechanism for Imbalance Classification in Network Digital Twins} 

\author[1,3]{Abubakar Isah} 

\author[1]{Ibrahim Aliyu}

\author[1]{Sulaiman Muhammad Rashid}

\author[1]{Jaehyung Park}

\author[4]{Minsoo Hahn}

\author[1]{Jinsul Kim\corref{cor1}} 
\ead{jsworld@jnu.ac.kr}
\cortext[cor1]{Corresponding authors}

\affiliation[1]{organization={Department of Intelligent Electronics and Computer Engineering, Chonnam National University},
            addressline={Buk-gu}, 
            postcode={61186}, 
            state={Gwangju},
            country={South Korea}}

\affiliation[2]{organization={ Electronics and Telecommunications Research Institute},
            postcode={34129}, 
            state={Daejeon},
            country={South Korea}}

\affiliation[3]{organization={Department of Computer Science, Ahmadu Bello University},
            postcode={810241}, 
            state={Zaria},
            country={Nigeria}}

\affiliation[4]{organization={Department of Computational and Data Science, Astana IT University, Astana, Kazakhstan}}



\newpageafter{abstract}
\begin{abstract}

Graph neural networks are gaining attention in fifth-generation (5G) core network digital twins, which are data-driven complex systems with numerous components. Analyzing these data can be challenging due to rare failure types, leading to imbalanced classification in multiclass settings. Digital twins of 5G networks increasingly employ graph classification as the main method for identifying failure types. However, the skewed distribution of failure occurrences is a significant class-imbalance problem that prevents practical graph data mining. Previous studies have not sufficiently addressed this complex problem. This paper, proposes class-Fourier GNN (CF-GNN) that introduces a class-oriented spectral filtering mechanism to ensure precise classification by estimating a unique spectral filter for each class. This work employs eigenvalue and eigenvector spectral filtering to capture and adapt to variations in minority classes, ensuring accurate class-specific feature discrimination, and adept at graph representation learning for complex local structures among neighbors in an end-to-end setting. The extensive experiments demonstrate that the proposed CF-GNN could help create new techniques for enhancing classifiers and investigate the characteristics of the multiclass imbalanced data in a network digital twin system.
 
\end{abstract}

\begin{graphicalabstract}
\includegraphics[width=1.0\textwidth]{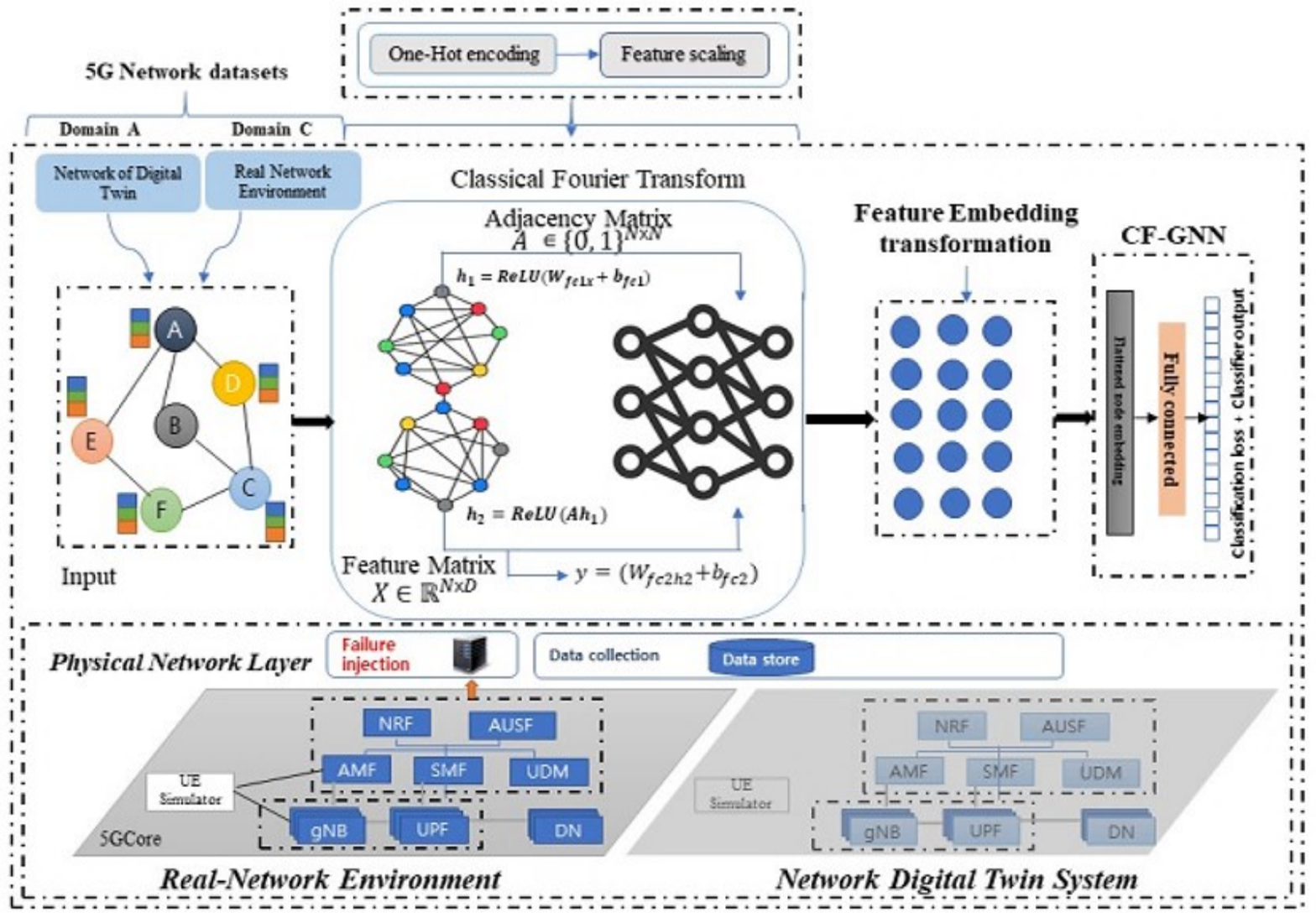}
\end{graphicalabstract}

\begin{highlights}
\item A novel CF-GNN addresses class-imbalance tasks in multiclass 5G network digital twins datasets.

\item Eigenvalues and eigenvectors detect the majority and minority classes of the failure types.

\item Global graph structures and local information affect spectral filtering.

\item The model performed robustly on the network digital twin-unbalanced datasets.
\end{highlights}


\end{frontmatter}



\section{Introduction} \label{Intro I}
The emergence of fifth-generation (5G) networks and robust core network systems is paramount for the evolution to Industry 5.0. According to recent studies, 5G connections are projected to increase more than 100 times, from approximately 13 million in 2018 to 2.6 billion by 2025 \cite{Ericsson:2025}. This growth is expected to continue with sixth-generation networks and the advent of the Industry 5.0 era, characterized by the vision of the Internet of Everything. Classifying network failures for 5G and beyond \cite{ramirez2020multilayer} is essential to optimize network performance. As the backbone of the communication infrastructure, reliable and resilient networks are vital for seamless operation in diverse applications in environments ranging from daily life to industrial settings \cite{isah2024graph}.

The stability, speed, and security of internet connectivity are critical to the reliable functioning of 5G networks \cite{rajak2024fdf}. However, disruptions in connectivity can lead to severe consequences. Given the interconnected nature of Internet Service Providers (ISPs), failures originating within a single ISP's domain can rapidly propagate across the entire network. As such, managing large-scale failures requires significant expertise. To mitigate the impact of such disruptions, rapid and automated anomaly detection is essential. The Border Gateway Protocol (BGP) \cite{fei2021analysis}, which interconnects the IP backbone of different ISPs, plays a pivotal role in this context. BGP routers must continuously update internal and external routing information to maintain service continuity. Consequently, the timely identification of hardware and software faults within these routers is crucial for sustaining 5G services \cite{Nokia:2025}. The growing complexity and volume of network traffic further complicate data-driven optimization strategies. In this setting, advanced AI techniques offer promising solutions by shifting the focus from reactive fault handling to proactive fault prediction, enabling preemptive interventions.

Graph classification has emerged as a solution to address the challenges of unbalanced graph classification in applications, including 5G network digital twins \cite{isahgft}, fraud detection \cite{liu2021pick, hu2023cost}, and, more recently, synthetic oversampling \cite{yan2025synthetic}, which has gained significant attention. Robust graph neural network (GNN) models must be developed to mitigate biases toward majority classes while ensuring effective generalization for minority classes. Traditional techniques for handling class imbalance, such as oversampling \cite{chawla2002smote} and classifier-based sampling \cite{hoens2013imbalanced}, have limitations when applied to graph-structured data because they do not consider the graph structure. Various methods have been proposed to address class imbalance within semi-supervised node classification.

Among the various approaches to address class imbalance in graph learning, synthetic oversampling techniques, including GraphSMOTE \cite{zhao2021graphsmote, zhao2024imbalanced} and GATE-GNN \cite{fofanah2024addressing}, generate new nodes and establish connections with the existing graph to improve class balance. Furthermore, the Boosting-GNN was introduced \cite{shi2021boosting} to address the classification of imbalanced nodes in graph networks, whereas another study \cite{hu2024gat} incorporated cost-sensitive learning into the framework. Studies have highlighted that Boosting-GNN improves computational efficiency by assigning higher weights to minority-class nodes during training, particularly in transferring learning applications.

Furthermore, reweighting techniques \cite{luo2025revive} and resampling methods \cite{avelino2024resampling} have been explored to mitigate class imbalance further in GNN-based node classification. In this context, loss functions have also been adapted based on class frequency and node degree, balancing the node distribution while preserving the underlying graph structure. Due to the context of addressing class imbalances in graph learning and focusing on how the loss function behaves in this frequency transformed space becomes critical. To this end, this work addresses the research question: \textit{(Q1) How does the consistency of the loss function variance across graphs within the whole graph affect class imbalance, particularly when spectral filtering transforms the graph structure from the time domain to the frequency domain?}

Nevertheless, failing to account for the class imbalance when designing GNN models can cause significant performance degradation. If the imbalance is not adequately addressed, the majority class tends to dominate the loss function, causing the trained GNN to overclassify majority class nodes while failing to predict minority class samples accurately. When applied to graphs with skewed class distributions, existing GNN methods frequently overfit the majority class, resulting in suboptimal embeddings for minority class nodes. Addressing this challenge is crucial for improving the generalizability of GNN models and ensuring their effectiveness beyond applications (e.g. 5G network digital twins). However, some studies have attempted to provide an interpretation of GNNs from a spectral perspective \cite{balcilar2021analyzing}, suggesting that these methods behave as low-pass filters. More recent work, including GNN based on node adaptive filtering \cite{zheng2023node, wu2024gnn}, extend this view by considering the influence of node degree, label distributions, and adaptive edge sampling mechanism on filter behaviours. Despite these advances, most GNNs are still primarily designed using a message-passing framework in the spatial domain, with limited exploration into how class-specific information influences spectral filter learning. This raises an important and underexplored question \textit{(Q2): How do individual nodes contribute to learning class-specific filter parameters in spectral GNNs? Are certain nodes given more weight during the filter parameter update process, particularly in imbalanced class settings?}

This study introduces a novel class-Fourier GNN (CF-GNN) filtering to address imbalance challenges in node-oriented classification and overcome these limitations. This approach employs class-specific spectral filtering, where distinct spectral filters are designed for each class instead of applying a uniform filter to all failure types. This approach ensures that minority-class nodes receive appropriate input without being overshadowed by majority-class nodes, enhancing their representation. The localized filtering mechanism highlights the distinct features of minority class nodes by focusing on specific node characteristics. This targeted approach enables class-specific transformations, ensuring these nodes are adequately represented during training and prediction. In addition, incorporating class-specific spectral awareness in the filtering process reduces the likelihood that minority class nodes are overwhelmed by dominant classes.

Furthermore, to improve classification performance, the proposed model employs class-weighted loss functions, which assign higher penalties to misclassifications of minority class samples. This approach encourages the network to learn more discriminative representations for underrepresented classes. The CF-GNN spectral filtering procedure explicitly integrates a multiclass structure, transforming node features in a class-aware manner using spectral coefficients and eigenvalue-based distinctions between classes. The proposed CF-GNN framework mitigates class-imbalance problems in graph learning, demonstrating substantial improvements over state-of-the-art GNN techniques in the experimental evaluations.

The contributions of this work and the proposed CF-GNN include the following:

\begin{itemize}
     \item The proposed CF-GNN employs a spectral filtering mechanism tailored to address class imbalance in multiclass node classification in network digital twins. Using the Fourier transform, CF-GNN captures intricate structural relationships that conventional spatial-based GNNs often neglect. This approach ensures that minority-class nodes are adequately represented, improving classification performance in imbalanced graph scenarios.
     
    \item The CF-GNN employs an iterative class-specific spectral filtering approach that dynamically refines eigenvector estimates of the graph structure. This process improves the representation of underrepresented classes by amplifying discriminative characteristics while mitigating the dominance of majority-class nodes in the embedding space, balancing the classification results. 

    \item We introduce the theoretical twin-graph Fourier transform (twin-GFT), a novel multidimensional spectral representation designed to enhance scalability and robustness in failure detection tasks, preserving critical frequency domain characteristics for various types of failures.
    
\end{itemize}

\subsection{Related work}
Graph learning-driven failure classifications have been gaining attention due to their effectiveness in addressing the massive data volume generated by 5G networks and their ability to understand their complex structural characteristics. This section explores two areas of research: learning-based 5G failure node prediction and imbalanced learning techniques, and the GFT to capture the complex structure of network digital twins.

\subsubsection{Learning-based failure prediction}

The primary advantage of applying graph deep learning to network modeling \cite{habibi2019comprehensive} is its data-driven approach, enabling it to capture the complex nature of the real-world networks. Most research \cite{ding2022data, said2020network} has employed fully connected conventional neural networks. However, the primary constraint of these methods is their non-generalizability to alternative network topologies and configurations, such as routing. In this context, more recent studies have proposed advanced neural network models, such as GNNs \cite{al2024comparative, soltanzadeh2021rcsmote}, convolutional neural networks \cite{said2020network}, and variational autoencoders \cite{ding2018opportunities}, for robustness in imbalanced datasets standard in cybersecurity domains. Despite their advancements, these models have different objectives and neglect critical aspects of real-world networks from the twin perspective. 

The concept of broadly applying neural networks to graphs has gained significant attention. For example, convolutional networks have been extended to graphs in the spectral domain \cite{bruna2013spectral}, where filters are applied to frequency modes using GFT. The eigenvector matrix of a graph Laplacian must be multiplied to achieve this transformation. One study \cite{defferrard2016convolutional} parameterized the spectrum filters as Chebyshev polynomials of eigenvalues, resulting in efficient and localized filters and reducing the computational burden. However, a disadvantageous drawback of these spectral formulations is that they are limited to graphs with a single structure because they depend on the fixed spectrum of the graph Laplacian.

In contrast, spatial formulations are not constrained by a specific graph topology. Generalizing neural networks to graphs is studied in two ways: a) given a single graph structure, or labels of individual nodes \cite{scarselli2008graph, bruna2013spectral, lei2017deriving, li2015gated, defferrard2016convolutional, kipf2016semi}, b) given a set of graphs with different structures and sizes. The goal is to learn predictions of the class labels of the graphs \cite{duvenaud2015convolutional, atwood2016diffusion, niepert2016learning}.  Inspired by the International Telecommunication Unit Challenge (ITU) "ITU-ML5G-PS-008: Network Failure Classification Model Using the Digital Twin Network" \cite{Junichi:2023}, this research aims to employ the enhanced CF-GNN to analyze interconnected systems and networks in an end-to-end manner. The proposed CF-GNN framework precisely classifies network failure types and accurately identifies failure points by integrating class-oriented spectral filtering. The approach improves accuracy and adaptability compared to conventional deep learning models, effectively capturing both global consistency and localized variations in the network structures.

\subsubsection{Imbalance learning technique}
\begin{figure}
    \centering
    \includegraphics[width=0.8\linewidth]{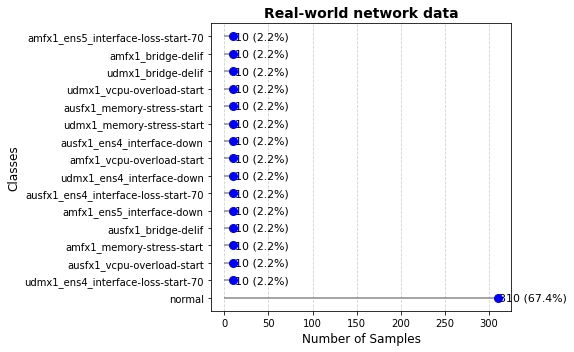}
    \caption{Class distribution of the failure types in real-world network data (Domain C).}
    \label{fig: real-network}
\end{figure}

\begin{figure}
    \centering
    \includegraphics[width=0.8\linewidth]{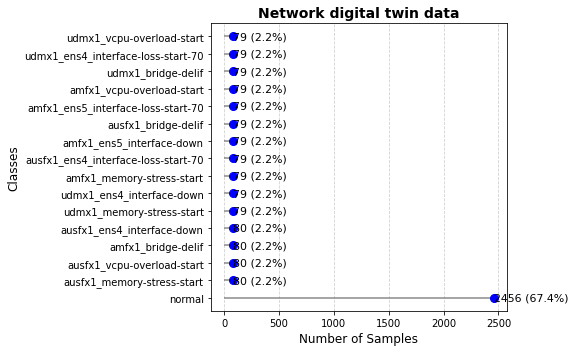}
    \caption{Class distribution of the failure types in network digital twins (Domain A).}
    \label{fig: NDT}
\end{figure}

    To investigate the underlying causes of failures in digital twin networks, the authors in \cite{kawasaki2020comparative, fei2021analysis} proposed a fault classification approach based on machine learning (ML). This method involves collecting 41 network-related attributes and employing ML algorithms to analyze these features to identify three primary root causes: CPU overload, node failure, and interface failure. For model evaluation, the study uses three algorithms: Multilayer Perceptron (MLP), Random Forest (RF), and Support Vector Machine (SVM) for failure detection. Figure \ref{fig: real-network} and Figure \ref{fig: NDT} represent the class distributions of the two datasets and the imbalance percentage of each failure type.
    Data-level and algorithm-level methods are the two main categories of class-imbalanced learning techniques. Before building classifiers, data-level approaches preprocess training data to reduce inequality \cite{li2021novel}. These strategies include undersampling the majority classes and oversampling the minority classes. In contrast, algorithm-level approaches address the problem of class imbalance by modifying the fundamental learning decision-making process of the model. Algorithm-level techniques can be broadly classified as threshold moving, cost-sensitive learning, and new loss functions \cite{johnson2019survey}. 
    
    Several approaches have been proposed at the data level. SMOTE \cite{chawla2002smote} generates synthetic minority samples in the feature space by interpolating existing minority samples and their nearest minority neighbors. However, one of the primary limitations of SMOTE is that it creates new synthetic samples without considering the neighborhood samples of the majority classes, which can increase class overlapping and noise \cite{koziarski2019radial}. Based on the SMOTE principle, many variations have been proposed (e.g.., Borderline-SMOTE \cite{han2005borderline} and Safe-Level-SMOTE \cite{bunkhumpornpat2009safe}), enhancing the original approach by considering majority class neighbors. Safe-Level-SMOTE creates safe zones to prevent oversampling in overlapping or noisy regions, whereas Borderline-SMOTE only samples the minority samples close to the class borders. One study \cite{gilmer2017neural} modified the parameters of the model learning process that favored classes with fewer samples to address class-imbalance challenges. For example, focal loss \cite{lin2017focal} allows minority samples to contribute more to the loss function. A novel loss function called mean squared false error \cite{wang2016training} was proposed to train deep neural networks on imbalanced datasets. The reviewed literature highlights significant findings from various studies; however, a key research gap remains unexplored and forms the basis of this study. Specifically, existing works have not thoroughly explored the selection of node failure points and optimal features to enhance classifier performance in graph data. Many prior studies concentrate solely on data from physical interfaces, overlooking a broader comparative evaluation of failure detection within network function virtualization (NFV) environments. Addressing this gap, the present work proposes a framework capable of distinguishing failure types from failure points and utilizing the most relevant features.

\subsection{Paper Structure}
This study enables us to handle the class-imbalanced graph data problem, and two new models have recently been developed. We employ some of the settings in GATE-GNN \cite{fofanah2024addressing} to dynamically modify the weights for the GNN modules and the Reweight \cite{he2024gradient} model, focusing on class-incremental learning in dynamic real-world environments regarding dual imbalances.
    
The paper is structured as follows: Section (\ref{2}) introduces the notation and preliminaries. Next Section (\ref{3}) investigates the graph filtering mechanism of the proposed method on network digital twin datasets. Then, Section (\ref{4}) presents the experiment and evaluation detailing the datasets and results. Finally, the work is concluded in Section (\ref{5}).

\begin{figure}
    \centering
    \includegraphics[width=0.7\linewidth]{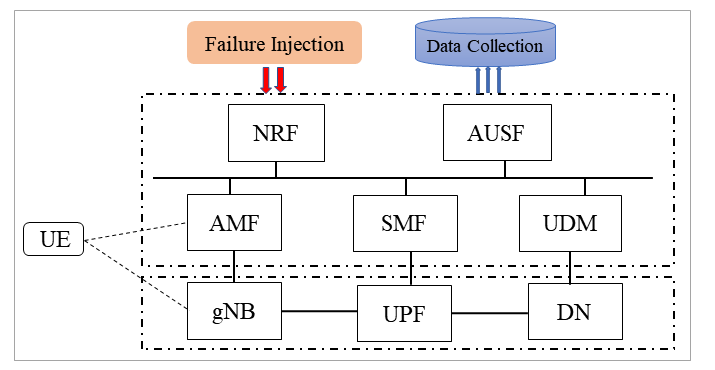}
    \caption{The components of fifth-generation core network digital twins failure injection settings.}
    \label{fig: 5G Core}
\end{figure}

\section{Notations and Preliminaries} \label{2}
We formulate the network failure injection task as a \textit{node classification} problem, aiming to learn expressive embeddings from rich operational data for accurate detection and differentiation of failure types. The underlying graph representation of the 5G core network is denoted as $G = (V, E)$, where $V$ is the set of network function nodes and $E$ represents the directed or undirected edges that capture operational dependencies and interactions between functions.

The adjacency matrix is defined as $A = \{ e_{vu} \mid \forall v, u \in V \} \in \mathbb{R}^{N \times N}$, which encodes the connectivity between nodes, where $e_{vu} = 1$ if an edge exists between nodes $v$ and $u$, and $e_{vu} = 0$ otherwise. Each node $v \in V$ corresponds to a specific core network function (e.g., AMF, AUSF, UDM) and is associated with a feature vector $f_v \in \mathbb{R}^{D}$, capturing operational metrics such as \texttt{packet counts, memory usage, CPU load, registration events, GTP traffic, and link speeds}. Collectively, the feature matrix is denoted as $F = \{ f_v \mid v \in V \} \in \mathbb{R}^{N \times D}$.

These features are derived from domain-specific categories including network interface statistics (e.g., \textit{out-octets}, \textit{unicast-pkts}), performance metrics (e.g., \textit{CPU usage}, \texttt{I/O wait}), user registration/authentication logs (e.g., \texttt{RegInitReq}, \texttt{RegInitSucc}), and traffic indicators (e.g., \textit{GTP.InDataPkt}, \textit{GTP.OutDataOct}). For instance, high \textit{out-unicast-pkts} values in AMF nodes may indicate control plane congestion, while changes in \textit{oper-status} in UPF nodes may signal tunnel instability.

Failures are assumed to manifest at the node level, and each node is assigned a class label $y_v \in Y$ indicating its operational state (e.g., \textit{normal}, \textit{failed}). Misclassifications are associated with node-specific costs based on criticality, with higher penalties for control-plane entities like AMF, AUSF, and UDM due to their impact on system reliability and service continuity.

\begin{figure}
    \centering
    \includegraphics[width=1.0\linewidth]{CF-GNN.pdf}
    \caption{The architectural framework of the proposed technique for imbalanced classification of the fifth-generation network of a digital twin}
    \label{fig: DTN 5G}
\end{figure}

 Figure \ref{fig: DTN 5G} illustrates the proposed model framework. We applied CF-GNN to examine individual nodes with various feature types (e.g., phys-address, oper-status, etc.). The GNN takes the provided network digital twin data as input and retrieves fine-grained information about node interactions and local graph structures. The class Fourier transform denoise features while capturing global network attributes. The graph module processes these inferred relationships to create a multilayer graph representation. The feature extraction applies the Fourier transform algorithm to convert the network data into useful information. The proposed CF-GNN integrates the extracted features with the learned graph structure to achieve accurate multiclass classification results.

\begin{table}[htbp]
\centering
\caption{Explanation of Graph and Laplacian Variables}
\label{tab:graph_variables}
\begin{tabular}{c|p{6cm}}
\hline
\textbf{Variable} & \textbf{Explanation} \\ 
\hline
$G = (V,E)$ & Graph representing the network digital twin system \\ 
\hline
$V_n$ & Set of nodes in graph $G_n$ \\ 
\hline
$N_n$ & Number of nodes in $G_n$ \\ 
\hline
$L_n$ & Laplacian matrix of $G_n$ \\ 
\hline
$\lambda^{(n)}_k$ & $k$-th eigenvalue of $\mathcal{L}^{(n)}$ \\ 
\hline
$\phi^{(n)}_k$ & Eigenfunctions of the Laplacian \\ 
\hline
$\mathcal{G}_1 \times \mathcal{G}_2$ & Product graph with node features \\ 
\hline
$\sigma(L_1 \oplus L_2)$ & Eigenvalues of $L_1 \oplus L_2$ \\ 
\hline
$(i_1, i_2)$ & Node indices in $\mathcal{G}_1$ and $\mathcal{G}_2$ \\ 
\hline
$(k_1, k_2)$ & Indices for eigenvalues and eigenfunctions \\
\hline
$y_{ij}$ & One-hot label component for node $v_i$ \\ 
\hline
$f_{ij}$ & Predicted probability for node $v_i$ in class $j$ \\ 
\hline
$h_i$ & Hidden representation at layer $i$ \\ 
\hline
$\hat{m}, \hat{v}$ & Adam optimizer moment estimates \\ 
\hline
$\eta$ & Learning rate \\ 
\hline
$y$ & Output vector of size $(N, C)$ \\ 
\hline
$\theta$ & Parameters of the GNN model \\ 
\hline
$f^{(l)}$ & Aggregation function at layer $l$ \\ 
\hline
$N(v_i)$ & Neighbors of node $v_i$ \\ 
\hline
$h_i^{(0)} = x_i$ & Initial feature vector for node $v_i$ \\ 
\hline
\hline
\end{tabular}
\end{table}

\subsection{ Graph neural network}
The GNN can transform the graph structure data into standard representations, making them suitable for input into neural training. We employed the concept of class spectral filtering introduced in \cite{yi2024fouriergnn} and \cite{guo2023graph} to enhance the learning process. This approach allows the GNN to propagate the information efficiently from the nodes and edges to their neighboring nodes. We use the neighbor aggregation approach in \cite{lessan2019hybrid} to achieve this propagation. The network infrastructure as depicted in Figure \ref{fig: 5G Core} is modeled as a graph $G = (V,E)$, where $V$ represents the node set (AMF, UDM, AUSF) and $E$ represents edge set. Each node $v_i$ in the graph is associated with a feature vector $x_i$, capturing relevant information regarding the corresponding network component. The graph can be represented using an adjacency matrix $A$, where $A_{ij} = 1$ if a connection exists between nodes $v_i$ and $v_j$, and $A_{ij} = 0$ otherwise. Node features are stored in the feature matrix $X$, where $X$ has the dimensions $|V| \times D$, where $|V|$ represents the number of nodes, and $D$ denotes the dimensionality of the feature vectors $X = [x_1, x_2, \ldots, x_{|V|}]$. 

Many network components depend upon one another due to their interconnectivity \cite{ferriol2023routenet}. The GNN architecture comprises of multiple layers, each processing information from neighboring vertices to extract hierarchical representations. We let $h_i^{(l)}$ denote the representation of vertex $v_i$ at layer $l$ of the GNN. Information is aggregated from the neighboring vertices and transformed using learnable parameters. The update equation for $h_i^{(l)}$ can be expressed as follows:

\begin{equation}
h_i^{(l)} = f^{(l)} \left( h_i^{(l-1)}, \{ h_j^{(l-1)} \}_{j \in N(v_i)} \right),
\end{equation}
where \( N(v_i) \) represents the set of neighboring vertices of \( v_i \).

One of the greatest challenges is creating network digital twin datasets that accurately replicate the real-world failure scenarios. This problem arises from several difficulties, including the limited availability of real-world system data and the inherent complexity of simulating various types of failure.

\subsection{Graph Fourier transform}
This work considers an undirected weighted graph \( G = (V, E, w) \), where \( V = \{0, 1, \ldots, N-1\} \) denotes the vertex set, \( E \) represents the edge set, and \( w(i,j) \) indicate the weight function satisfying \( w(i,j) = 0 \) for any \( (i,j) \) $\notin$ \( E \). This graph is assumed to be simple,  meaning it contains no self-loops. Each node in this graph corresponds to a functional entity in the 5G core architecture, and edges represent measurable interactions (e.g., packet transfer, registration signaling, GTP flows). These matrices associated with \( G \) are important: the adjacency matrix \( W \), a degree matrix \( D \), and a Laplacian matrix \( L = D-W \). The Laplacian matrix is essential for GFTs.

The Laplacian matrix \( L \) is real, symmetric, and positive-semidefinite, thus, it possesses nonnegative eigenvalues \( \lambda_0, \ldots, \lambda_{N-1} \) and the corresponding orthonormal eigenfunctions \( u_0, \ldots, u_{N-1} \). These eigenfunctions satisfy the following equation:

\begin{equation}
L \begin{bmatrix}
u_k(0) \\
\vdots \\
u_k(N-1)
\end{bmatrix} = \lambda_k \begin{bmatrix}
u_k(0) \\
\vdots \\
u_k(N-1)
\end{bmatrix}
\end{equation}

For \( k = 0, \ldots, N-1 \), orthonormality implies that the sum of the products of the corresponding eigenfunctions equal the Kronecker delta function \(\delta(i,j)\), where \(\delta(i,j)\) equals 1 if \(i = j\) and is zero otherwise. This study assumes that eigenvalues are arranged in ascending order, such that \(\lambda_0 \leq \ldots \leq \lambda_{N-1}\). In addition, \(\lambda_0\) is strictly zero because the sum of rows in \(L\) equals zero. The matrix spectrum, denoted by \(\{ \lambda_k \}_{k=0}^{N-1}\), is represented as \(\sigma(L)\).
The GFT in \cite{kurokawa2017multi} of a graph feature \( f : V \rightarrow \mathbb{R} \) is defined as \( \hat{f} : \sigma(L) \rightarrow \mathbb{C} \), where \( \sigma(L) \) represents the spectrum of the graph Laplacian \( L \). It is expressed as follows:

\begin{equation}
    \hat{f}(\lambda_k) = \langle f, u_k \rangle = \sum_{i=0}^{N-1} f(i) u_k(i),
\end{equation}

for \( k = 0, \ldots, N-1 \), where, \( u_k \) denotes the orthonormal eigenfunctions of \( L \), and the GFT represents a feature expansion using these eigenfunctions. The inverse GFT is given by,

\begin{equation}
    f(i) = \sum_{k=0}^{N-1} \hat{f}(\lambda_k) u_k(i),
\end{equation}

which reconstructs the original function $f(i)$ from its spectral components. The basic function of the GFT in spectral graph research is highlighted because it is equal to the discrete Fourier transform on cycle graphs.

When a graph Laplacian has non-distinct eigenvalues, the functions generated by the GFT may not be well-defined, resulting in multi-valued functions. For instance, if two orthonormal eigenfunctions, \( u \) and \( u_0 \) correspond to the same eigenvalue \( \lambda \), then the spectral component \( \hat{f}(\lambda) \) can have two distinct values: \( \langle f,u \rangle \) and \( \langle f,u_0 \rangle \).

\subsection{Digital twin spectral filtering}

The Cartesian product \( G_1 \square G_2 \) of graphs \( G_1 = (V_1, E_1, w_1) \) and \( G_2 = (V_2, E_2, w_2) \) is a graph with the vertex set \( V_1 \times V_2 \) and the edge set \( E \) defined as follows. For any \((i_1, i_2)\) and \((j_1, j_2)\) in the vertex set \( V_1 \times V_2 \), these are connected by an edge if either \((i_1, j_1)\) is in \( E_1 \) and \(i_2 = j_2\) or \(i_1 = j_1\) and \((i_2, j_2)\) is in \( E_2 \). The weight function \( w \) is defined as follows:

\begin{equation}
    w((i_1, i_2), (j_1, j_2)) = w_1(i_1, j_1) \delta(i_2, j_2) + \delta(i_1, j_1) w_2(i_2, j_2)
\end{equation}

where \( \delta \) denotes the Kronecker delta function. The graphs \( G_1 \) and \( G_2 \) are referred to as the factor graphs of \( G_1 \square G_2 \).

The adjacency, degree, and Laplacian matrices of the Cartesian product graph can be derived from those of its factor graphs \cite{zheng2015redundant}. Two factor graphs, \( G_1 \) and \( G_2 \), with vertex sets \( V_1 = \{0, 1, \ldots, N_1 - 1\} \) and \( V_2 = \{0, 1, \ldots, N_2 - 1\} \) respectively are considered. Each with adjacency matrix \( W_1 \) and \( W_2 \), degree matrix \( D_1 \) and \( D_2 \), and Laplacian matrix \( L_1 \) and \( L_2 \) respectively. When the vertices of the Cartesian product graph are ordered lexicographically, such as \((0, 0), (0, 1), (0, 2), \ldots, (N_1 - 1, N_2 - 1)\), the adjacency, degree, and Laplacian matrices of \( G_1 \square G_2 \) can be expressed as \( W_1 \oplus W_2 \), \( D_1 \oplus D_2 \), and \( L_1 \oplus L_2 \), respectively, where the operator \( \oplus \) denotes the Kronecker sum.

\textbf{Definition 1:} The Kronecker sum is defined by \( A \oplus B = A \otimes I_n + I_m \otimes B \) for matrices \( A \in \mathbb{R}^{m \times m} \) and \( B \in \mathbb{R}^{n \times n} \), where \( I_n \) represents an identity matrix of size \( n \).

The Kronecker sum has a valuable characteristic that decomposes an eigenproblem involving the Laplacian matrix of a product graph into eigenproblems of Laplacian matrices of the factor graphs. We assumed that the Laplacian matrix \( L_n \) of each factor graph has nonnegative eigenvalues \(\{\lambda^{(n)}_k\}_{k=0}^{N_n-1}\) and orthonormal eigenfunctions \(\{u^{(n)}_k\}_{k=0}^{N_n-1}\) for \(n=1,2\). In this case, the Kronecker sum \( L_1 \oplus L_2 \) yields an eigenvalue of \(\lambda^{(1)}_k + \lambda^{(2)}_k\) and the corresponding eigenfunction \(u^{(1)}_k \otimes u^{(2)}_k : V_1 \times V_2 \rightarrow \mathbb{C}\), where \(\oplus\) denotes the Kronecker sum. This eigenfunction satisfies the following:

\begin{align}
& (L_1 \oplus L_2) \left[ \begin{array}{c} u^{(1)}_k(0) u^{(2)}_k(0) \\ u^{(1)}_k(0) u^{(2)}_k(1) \\ \vdots \\ u^{(1)}_k(N_1-1) u^{(2)}_k(N_2-1) \end{array} \right] \\
& = (\lambda^{(1)}_k + \lambda^{(2)}_k) \left[ \begin{array}{c} u^{(1)}_k(0) u^{(2)}_k(0) \\ u^{(1)}_k(0) u^{(2)}_k(1) \\ \vdots \\ u^{(1)}_k(N_1-1) u^{(2)}_k(N_2-1) \end{array} \right]
\end{align}

This property holds for any \(k_1 = 0, \ldots, N_1-1\) and \(k_2 = 0, \ldots, N_2-1\). The resulting eigenvalues \(\{\lambda^{(1)}_k + \lambda^{(2)}_k\}\) are nonnegative, and the corresponding eigenfunctions \(\{u^{(1)}_k \otimes u^{(2)}_k\}\) are orthonormal. These conclusions can be directly derived from the fundamental properties of the Kronecker product.

A similar approach applies to decomposing an eigenproblem related to an adjacency matrix of a product graph. If the adjacency matrix $W_n$ has eigenvalues $\{\mu^{(n)}_k\}_{k=0}^{N_n-1}$ and orthonormal eigenfunctions $\{\mathbf{v}^{(n)}_k\}_{k=0}^{N_n-1}$ for $n=1,2$, the Kronecker sum $W_1 \oplus W_2$ yields the eigenvalues $\{\mu^{(1)}_k + \mu^{(2)}_l\}$ and corresponding eigenfunctions $\{\mathbf{v}^{(1)}_k \otimes \mathbf{v}^{(2)}_l\}$ defined on $V_1 \times V_2$.

The Cartesian product graph is represented as $G_1 \times G_2$, where $G_n$ for $n=1,2, G_n$ is an undirected weighted graph with a vertex set $V_n = \{0,1,\dots,N_n-1\}$. We assume its graph Laplacian $L_n$ has eigenvalues $\{\lambda_k^{(n)}\}_{k=0}^{N_n-1}$ and corresponding orthonormal eigenfunctions $\{u_k^{(n)}\}_{k=0}^{N_n-1}$. The GFT of a graph feature $f:V_1 \times V_2 \rightarrow \mathbb{R}$ on the product graph $G_1 \times G_2$ is represented by,

\begin{equation}
    \hat{f} : (\sigma(L_1) \otimes \sigma(L_2)) \rightarrow \mathbb{C}
\end{equation}

\begin{equation} \label{Eq:9}
\hat{f}(\lambda^{(1)}_{k_1} + \lambda^{(2)}_{k_2}) = \sum_{i_1=0}^{N_1-1} \sum_{i_2=0}^{N_2-1} f(i_1,i_2) \cdot u^{(1)}_{k_1}(i_1) \cdot u^{(2)}_{k_2}(i_2)
\end{equation}

for $k_1=0,\dots,N_1-1$ and $k_2=0,\dots,N_2-1$, and its inverse is:

\begin{equation}\label{Eq:10}
f(i_1,i_2) = \sum_{k_1=0}^{N_1-1} \sum_{k_2=0}^{N_2-1} \hat{f}(\lambda^{(1)}_{k_1} + \lambda^{(2)}_{k_2}) \cdot u^{(1)}_{k_1}(i_1) \cdot u^{(2)}_{k_2}(i_2)
\end{equation}

Thus, this work interprets features on a Cartesian product graph as two-dimensional features and proposes a twin GFT, enabling multiclass failure classification in Network Digital Twin environments.

\textbf{Definition 1} The twin GFT of a function $f: V_1\times V_2\rightarrow \mathbb{R}$ on a Cartesian product graph $G_1 \square G_2$ is a spectral representation $\hat{f}:\sigma(L_1)\times \sigma(L_2) \rightarrow \mathbb{C}$ defined as follows:

\begin{equation} \label{Eq:11}
    \hat{f}(\lambda_{11},\lambda_{22}) = \sum_{i_1=0}^{N_1-1} \sum_{i_2=0}^{N_2-1} f(i_1,i_2) \cdot u_{k_{11}}^{(1)}(i_1) \cdot u_{k_{22}}^{(2)}(i_2)
\end{equation}

for $k_{11}=0,\dots,N_1-1$ and $k_{22}=0,\dots,N_2-1$, and its inverse is given by:

\begin{equation} \label{Eq:12}
    f(i_1,i_2) = \sum_{k_1=0}^{N_1-1} \sum_{k_2=0}^{N_2-1} \hat{f}(\lambda_{k_{11}}^{(1)} + \lambda_{k_{22}}^{(2)}) \cdot u_{k_{11}}^{(1)}(i_1) \cdot u_{k_{22}}^{(2)}(i_2)
\end{equation}

for $i_1=0,\dots,N_1-1$ and $i_2=0,\dots,N_2-1$.

The twin GFT can be represented as a series of matrix-to-matrix multiplications. The twin GFT applied to features f is expressed using $N_1 \times N_2$ matrices $F=(f(i_1,i_2))_{i_1,i_2}$ and $\hat{F}=(\hat{f}(\lambda_{k_{11}}^{(1)} + \lambda_{k_{22}}^{(2)}))_{k_{11},k_{22}}$, as follows:

\begin{equation} \label{Eq:13}
\hat{F} = U_1^* FU_2,
\end{equation}

where $U_n$ denotes an $N_n \times N_n$ unitary matrix with the $(i,k)$-th element $u_{k}^{(n)}(i)$ for $n=1,2$. Then, its inverse transform is given as follows:

\begin{equation} \label{Eq:14}
    F = U_1 \hat{F}^2 U_2^*,
\end{equation}

The twin GFT has connections to existing transformations. First, when both factor graphs are cycle graphs, the twin GFT can be equivalent to the 2D-Discrete Fourier Transform and the 2D-discrete Fourier transform and 2D GFT. Second, when a factor graph is a cycle graph, some cycles might be nested within other cycles, creating a complex network structure. The twin GFT is a joint graph and temporal Fourier transform, generalizing the existing transformations.

\section{Proposed Class-Fourier Graph Neural Network} \label{3}
The CF-GNN performs well with homophilic and heterophilic graphs and has demonstrated good expressive capacities. However, the current methods often lack adaptability. Conventional approaches apply a single, globally shared spectral filter $\hat{g}$, trained across the entire graph, with fixed frequency coefficients $\{\gamma_k\}_{k=0}^K$. Node-specific changes are not considered in the global filter, which cannot distinguish between the distinct local structures connected to each node during the filtering process. Although polynomial-parameterized spectral filter learning offers a certain level of localization, it might not adequately capture subtle and diverse local structural patterns. It is intuitively reasonable to train a class-specific filter $\hat{g}_i(\lambda_l)$ rather than to depend on a globally consistent spectral filter $\hat{g}(\lambda_l)$. A more efficient method is provided by $\hat{g}_i(\lambda_l)$, tailored for each node $i$ to adaptively capture its local patterns adaptively. This section revisits globally consistent spectral graph transforms and proposes a spectral filter learning strategy in the context of CF-GNNs to mitigate this limitation.

\subsection{Class-Fourier transform filtering}

This section introduces an adaptive localized spectral filtering method for CFT filtering on a graph \( G \), inspired by the generalized translation operator from graph signal processing (GSP) in \cite{zheng2023node}. Using polynomial-parameterized spectral filtering as described in \cite{guo2024rethinking}, this approach ensures accurate and flexible representation of local patterns by incorporating the influence of the node at which the filter is applied.

We define a generalized translation operator \( T_i \) for any signal \( g \in \mathbb{R}^n \) defined on a graph \( G \), and for any node \( i \in \{0, 1, \dots, n-1\} \), as:

\begin{equation} \label{Eq:15}
    T_i : \mathbb{R}^n \to \mathbb{R}^n,
\end{equation}

and specifically,

\begin{equation} \label{Eq:16}
    T_i(g) := \sqrt{N} \, (g \ast \delta_i) = \sqrt{N} \sum_{l=1}^{n} u_l u_l^\top(i) \hat{g}(\lambda_l),
\end{equation}

where \( u_l^\top(i) \) denotes the \( i \)-th entry of the eigenvector \( u_l \) corresponding to eigenvalue \( \lambda_l \), and \( \hat{g}(\lambda_l) \) is the spectral representation of signal \( g \).

In the context of CFT filtering, adaptive local filtering is achieved by centering the filter signal \( g \) at the target node \( v_i \) via \( T_i \), followed by spectral convolution with \( x \):

\begin{equation} \label{Eq:17}
    x \ast G T_i(g) = \sum_{l=1}^N u_l \, x^{(\lambda_l)} u_l^\top(i) g^{(\lambda_l)}
\end{equation}

Since \( \hat{g}_i(\lambda_l) = \sqrt{N} \, u_l(i) \, \hat{g}(\lambda_l) \), we can rewrite Equation \ref{Eq:17} as:

\begin{equation} \label{Eq:18}
     x \ast G T_i(g) = \sum_{l=1}^N u_l \, x^{(\lambda_l)} g_i^{(\lambda_l)}
\end{equation}

Noting that \( x_i = U_{i:} \cdot \hat{x} \), where \( U_{i:} \) is the \( i \)-th row of the eigenvector matrix \( U \), and \( q = \text{pinv}(\hat{x}) \in \mathbb{R}^n \), we approximate:

\begin{equation} \label{Eq:19}
    U_{i:} \approx x_i q, \quad \hat{g}_i(\lambda_l) \approx \tilde{g}_i(\lambda_l) = \sqrt{N} (x_i q_l) \hat{g}(\lambda_l)
\end{equation}

with \( q_l \) the \( l \)-th element of \( q \).

This leads to a node-specific filter \( \tilde{g}_i \) that reflects the influence of feature \( x_i \). The filter can be parameterized using a \( K \)-order polynomial approximation:

\begin{equation} \label{Eq:20}
    \tilde{g}_i = \sum_{k=0}^{K} \eta_{i,k} \hat{p}_k(\Lambda),
\end{equation}

where \( \eta_{i,k} \) are trainable coefficients. Focusing on the filter output at each node, we compute:

\begin{equation} \label{Eq:21}
    z_i = \delta_i (U \tilde{g}_i U^\top x) = \delta_i U \sum_{k=0}^{K} \eta_{i,k} \hat{p}_k(\Lambda) U^\top x
\end{equation}

\begin{equation} \label{Eq:22}
    = \delta_i \sum_{k=0}^{K} \Psi_{i,k} \hat{p}_k(L)x
\end{equation}

with \( \Psi = [\eta_{i,k}] \in \mathbb{R}^{n \times (K+1)} \). This extends to a feature matrix \( X \) as:

\begin{equation} \label{Eq:23}
    Z_i = \delta_i \sum_{k=0}^{K} \Psi_{i,k} \hat{p}_k(L) X
\end{equation}

\subsubsection*{CF-GNN-V: Eigenvalue-based Transformation}
This variant applies a transformation to the learned features in the spectral domain that emphasizes eigenvalue-related frequency components relevant to the minority class. It introduces a class-sensitive weighting function \( \gamma(\lambda) \) such that:

\begin{equation} \label{Eq:24}
    Z_i = \delta_i \sum_{k=0}^{K} \Psi_{i,k} \gamma(\Lambda) \hat{p}_k(L) X
\end{equation}

Here, \( \gamma(\Lambda) \) is a diagonal matrix scaling frequency components to highlight class-specific spectral patterns.

\paragraph{Proof of CF-GNN\textsubscript{v}}
The base form of graph convolution in the spectral domain is:
\begin{equation}
    Z_i = \delta_i \sum_{k=0}^{K} \Psi_{i,k} \hat{p}_k(L) X
\end{equation}

This corresponds to filtering the signal \( X \) using polynomial spectral filters \( \hat{p}_k(\Lambda) \), followed by node-wise localization using \( \delta_i \).

In CF-GNN\textsubscript{v}, we introduce a class-sensitive transformation \( \gamma(\Lambda) \) into the filtering process:

\begin{equation} \label{Eq:25}
    Z_i = \delta_i \sum_{k=0}^{K} \Psi_{i,k} \gamma(\Lambda) \hat{p}_k(L) X
\end{equation}

In spectral graph theory, filtering a signal \( x \in \mathbb{R}^n \) is represented as:
\begin{equation} \label{Eq:26}
    x' = U g(\Lambda) U^\top x
\end{equation}

where \( g(\Lambda) \) is a diagonal matrix that applies the filter in the spectral domain.

To emphasize frequency components relevant to the minority class, we define \( \gamma(\lambda) \) to selectively modulate these components. Since \( \gamma(\Lambda) \) and \( \hat{p}_k(\Lambda) \) are diagonal, we can define a composite filter:

\begin{equation} \label{Eq:27}
    \tilde{g}_{i,k}(\Lambda) = \gamma(\Lambda) \hat{p}_k(\Lambda)
\end{equation}

which leads to:
\[
    Z_i = \delta_i \sum_{k=0}^{K} \Psi_{i,k} U \tilde{g}_{i,k}(\Lambda) U^\top X
\]

\[
= \delta_i \sum_{k=0}^{K} \Psi_{i,k} U \gamma(\Lambda) \hat{p}_k(\Lambda) U^\top X
\]

\begin{equation} \label{Eq:28}
    = \delta_i \sum_{k=0}^{K} \Psi_{i,k} \gamma(\Lambda) \hat{p}_k(L) X
\end{equation}

This final expression shows that CF-GNN-V modifies the spectral filter to reflect class-sensitive frequency emphasis.

\subsubsection{Eigenvector-aware attention}
We begin with the classical spectral graph convolution expression, as shown earlier in Eq.~\eqref{Eq:26}:

\begin{equation}
x' = U g(\Lambda) U^\top x
\end{equation}

Now, instead of only manipulating the eigenvalues $\Lambda$ using $g(\Lambda)$, we propose to modify the spectral projection basis $U$ using eigenvector-aware attention $\alpha(U)$, where $\alpha(U) \in \mathbb{R}^{n \times n}$ could be either a full matrix or a diagonal one (simpler to implement and more stable in training).

The modified filtering of the signal is:

\begin{equation} \label{Eq:30}
x' = U \alpha(U) g(\Lambda) U^\top x
\end{equation}

This formulation applies an attention mechanism in the spectral domain on the eigenvectors, rather than just adjusting the spectral coefficients (eigenvalues). The idea is that certain eigenvectors capture structural features (e.g., localized clusters, boundary conditions) that may be more informative for classifying failure modes, especially in minority classes.

We define the composite filter with eigenvector attention as:

\begin{equation} \label{Eq:31}
\tilde{g}_{i,k}(U, \Lambda) = \alpha(U) \hat{p}_k(\Lambda)
\end{equation}

Substituting into the spectral convolution yields:

\begin{equation} \label{Eq:32}
Z_i = \delta_i \sum_{k=0}^{K} \Psi_{i,k} U \alpha(U) \hat{p}_k(\Lambda) U^\top X 
= \delta_i \sum_{k=0}^{K} \Psi_{i,k} \alpha(U) \hat{p}_k(L) X
\end{equation}

This final expression shows that CF-GNN\textsubscript{E} introduces class-aware attention over eigenvectors in the spectral filtering process, complementing the eigenvalue-focused approach of CF-GNN\textsubscript{v}.

\begin{algorithm}[H]
\caption{CF-GNN: Class-Fourier Graph Neural Network}
\KwIn{Symmetric matrix $\mathbf{A} \in \mathbb{R}^{n \times n}$}
\KwIn{Eigenvalues $\lambda_i$ and eigenvectors $\mathbf{v}_i$}
\KwOut{Classification result $C(v), v \in V_{\text{train}}$}

// \textit{Initialization}

Set initial eigenvector guess: $\mathbf{v}^{(0)} \gets \text{random unit vector}$ \\
Set tolerance $\epsilon > 0$ \\
Set iteration count $t \gets 0$ \\

// \textit{Train CF-GNN}
\For{$l = 1, 2, \dots, L$}{  
    \For{$e = 1, 2, \dots, E$}{  
        // \textbf{GNN training}  \\
        $h_v^{(l)} \gets$ \text {Fourier transform on node embeddings};  \\
        $\mathcal{L}_{\text{GNN}}$ using spectral convolution;  \\
        $\mathcal{F}(v)$ Compute Fourier coefficients;  \\
        
        // \textbf{Feature embedding transformation}  \\
        $X^{(l)} \gets$ Transformed Fourier domain features;  \\
        $p_k^{(l)}(v) \gets$ Compute probability vector;  \\
        Calculate feature update loss $\mathcal{L}_{\text{MLP}}$;  \\
        Compute overall loss of CF-GNN $\mathcal{L}_{\text{CF-GNN}}$;  \\
        
        // \textbf{Eigenvalue and eigenvector computation} \\
        \While{not converged}{
            Compute $\mathbf{w}^{(t)} \gets \mathbf{A} \mathbf{v}^{(t)}$ \\
            Compute eigenvalue estimate: $\lambda^{(t)} \gets \frac{\mathbf{w}^{(t)} \cdot \mathbf{v}^{(t)}}{\|\mathbf{v}^{(t)}\|}$ \\
            Normalize eigenvector: $\mathbf{v}^{(t+1)} \gets \frac{\mathbf{w}^{(t)}}{\|\mathbf{w}^{(t)}\|}$ \\
            Check convergence: \textbf{if} $\|\mathbf{v}^{(t+1)} - \mathbf{v}^{(t)}\| < \epsilon$ \textbf{then stop} \\
            Update iteration count: $t \gets t + 1$
        }
        
        // \textbf{Class-specific spectral filtering} \\
        For each node $v_i$, \\
        compute Eq (\ref{Eq:18}) spectral filter $\tilde{g}_i(\lambda_l)$
        
        // \textbf{Polynomial approximation of localized filter} \\
        Approximate $\tilde{g}_i$ using a $K$-order polynomial in Eq(\ref{Eq:19}):        
        // \textbf{Class-oriented filtering} \\
        Compute filtered output (\ref{Eq:20}) for node $v_i$:
    }
}
\textbf{Ensemble classification result} $H(v) \gets$ from all weak classifiers;  

\end{algorithm}

\subsection{Experimental settings} \label{Experiment}

\subsubsection{Description of the KDDI datasets} 
The KDDI dataset is based on the ITU Challenge "ITU-ML5G-PS-008: Network Failure Classification Model Using Network Digital Twin" \cite{Junichi:2023}. The dataset comprises (Domain A) derived from the simulated environment of a network digital twin, and (Domain C) a dataset generated from a real-world network. Each domain dataset comprises 4,121 features, with 16 failure classes per failure sample. However, failure-type classes consist of a “normal” class and 15 other classes, with more than 67\% of the data belonging to the “normal” class. In comparison, the remaining 15 failure classes comprise only 2.2\% of the dataset. Table~\ref{tab:a1} presents the class distribution for the two datasets, whereas Table~\ref{tab:failure_classes} lists each failure point along with its corresponding failure types, some features, and descriptions relevant to the datasets used.

\renewcommand{\arraystretch}{0.9} 
\begin{table}[htbp]
\centering
\caption{Description of the dataset based on failure nodes, failure points, classes, and some features used for the classification.}
\scriptsize
\setlength{\tabcolsep}{3pt} 
\begin{tabular}{lllp{3.7cm}p{3.9cm}}
\toprule
\textbf{Node} & \textbf{Types} & \textbf{Classes} & \textbf{Features} & \textbf{Description} \\
\midrule

\multirow{5}{*}{AMF} 
& \multirow{5}{*}{amfx1} 
& bridge-delif & amfx1\_statistics.out-octets & \multirow{5}{=}{Tracks AMF packet flow for mobility. Monitors CPU/memory load to identify failure. } \\
& & ens5 inter-down & out-unicast-pkts & \\
& & ens5 interface-loss & memory-stats.used-percent\_value & \\
& & memory-stress-start & per-core-CPU-system\_value & \\
& & vcpu-overload-start & & \\

\midrule

\multirow{5}{*}{AUSF} 
& \multirow{5}{*}{ausfx1} 
& bridge-delif & RM.RegisteredSubNbrMax & \multirow{5}{=}{Measures registered subscribers for authentication. Monitors GTP traffic from AUSF} \\
& & inter-down & RM.RegisteredSubNbrMean & \\
& & inter-loss-70 & GTP.OutDataOctN3UPF & \\
& & memory-stress-start & GTP.OutDataPktN3UPF & \\
& & vcpu-overload-start & & \\

\midrule

\multirow{5}{*}{UDM} 
& \multirow{5}{*}{udmx1} 
& bridge-delif & udmx1\_five-g.RM.RegInitReq & \multirow{5}{=}{Handles user profile access and registration. Tracks traffic for UDM data management.} \\
& & inter-down & udmx1\_ens5\_statistics.out-unicast-pkts\_value & \\
& & inter-loss-70 & udmx1\_ens5\_speed\_value & \\
& & memory-stress-start & ens5.out-unicast-pkts & \\
& & vcpu-overload-start & ens5.speed & \\

\bottomrule
\end{tabular}
\label{tab:failure_classes}
\end{table}

\subsubsection{Parameter settings}
We randomly select the training samples while maintaining the same positive-to-negative sample ratio as in the entire dataset. The Adam optimizer is employed for parameter optimization, and the hyperparameters are configured as follows: For the network digital twin dataset (Domain A), the hidden embedding size is set to 64 with a \( L_2 \) regularization value of \( 5e{-4}\), the learning rate to 0.01, the adjacency dropout to 0.2, with two model layers, and training is conducted for 350 epochs. For the real-world network dataset (Domain C), the hidden embedding size is also set to 64, weight loss $(\lambda)$ is 1e-6, learning rate to 0.01, the number of model layers to 2, and the maximum number of epochs to 250, and imbalance ratio is set 0.1 to 0.9. The proposed method is implemented using PyTorch and is executed in a Python 3.8 environment on a Windows 10 machine equipped with a 64-bit operating system, a 13th-generation Intel Core i5-13400 processor, 32 GB of RAM, and an NVIDIA GeForce RTX 3060 Ti GPU.

\subsubsection{Preprocessing}
To prepare the data for input into the GNN model, we performed two primary preprocessing steps: categorical encoding and feature normalization.
Categorical attributes are transformed into binary vectors via one-hot encoding. Given a categorical attribute with $k$ possible values, we create a binary vector of length $k$, where only the index corresponding to the actual category is set to 1. For instance, Figure \ref{fig: 5G Core} represents the components of the three failure point functions (AMF, AUSF, and UDM) and is encoded as $[1, 0, 0]$, $[0, 1, 0]$, or $[0, 0, 1]$, respectively.

To ensure scale invariance and stabilize training, we apply min-max normalization \cite{ieracitano2020novel} to all numeric features. The transformation is defined by:

\begin{equation}
f_A(x_i) = \tilde{x}_i = \frac{x_i - \min(x_j)}{\max(x_j) - \min(x_j)},
\end{equation}

where $x_i$ is the $i^{\text{th}}$ value of the numeric attribute $A$, and $\min(x_j)$, $\max(x_j)$ denote the minimum and maximum values in attribute $A$, respectively. This mapping scales all values into the range $[0, 1]$, which improves convergence and comparability between different types of features.

\begin{table}[h]
    \centering
    \caption{Dataset distribution table}
    \label{tab:a1}
    \begin{tabular}{c|c|c|c}
        \hline
        \textbf{Dataset} & \textbf{Failure events} & \textbf{Features} &\textbf{IR} \\
        \hline
        \multirow{2}{*}{Domain A } & Normal = 2456 
        &\multirow{2}{*}{4,121} 
        &\multirow{2}{*}{0.4886} \\
        & Failure = 1186 & \\
        \hline
        \multirow{2}{*}{Domain C } & Normal = 588 
        &\multirow{2}{*}{4,121} 
        &\multirow{2}{*}{0.4847} \\
        & Failure = 285 & \\
        \hline
       
    \end{tabular}
\end{table}

\subsection{Comparison with the baseline methods}
The CF-GNN presented in this article is a GNN model designed to identify failure classes in a network digital twin setting. The primary scientific problem addressed is graph representation learning for imbalanced graphs. We compared the results with innovative baseline techniques that include: 1) general GNN models, GCN \cite{kipf2016semi}, and GAT \cite{velickovic2017graph}; and 2) four imbalance-based models, including PC-GNN \cite{liu2021pick}, Reweight \cite{he2024gradient}, GraphSMOTE \cite{zhao2021graphsmote}, and GATE-GNN \cite{fofanah2024addressing}, as state-of-art GNN imbalance classification models.

\begin{itemize}
    
    \item \textit{GCN}: The GCN aggregates information and updates its data presentation from the neighboring nodes, repeating this process across several layers.
    \item \textit{GAT}: The GAT offers more accurate representations of the graph than earlier GNNs by enabling the network to learn which neighbors are the most significant for each node.
    \item \textit{In Reweight Loss} the core idea of gradient reweighting is to mitigate the effects of class imbalance by adjusting the gradient updates during training.
    \item \textit{GraphSMOTE} technique generates synthetic nodes while preserving the graph structure to address the challenge of imbalanced node classification in graphs.
    \item \textit{PC-GNN} is exposed to a more balanced training input by sampling balanced subgraphs, maintaining the domination of the majority classes in the learning process.
    \item \textit{GATE-GNN} method for a classification of imbalanced nodes, creates dynamic node interactions within the GATE architecture using learnable weight parameters.
    \item In the {CF-GNN}, the proposed model improves performance, particularly for minority classes in imbalanced graph classification, by employing the Fourier transform to examine class-specific frequency features.
    \item {CF-GNN\_v} applies a spectral-inspired transformation to amplify frequency components associated with class distinctions.
    \item {CF-GNN\_E}: Uses attention over spectral embeddings to focus on structurally informative nodes. Boosts sensitivity to underrepresented class patterns in imbalanced graph data.
\end{itemize}

\subsection{Evaluation Metrics}
The selection of evaluation metrics is essential when addressing imbalanced datasets. These metrics must provide a fair assessment of classification results across all classes and pay special attention to the performance of minority classes \cite{isahgft}. We chose five widely used metrics to ensure an objective evaluation: the precision, recall, F1-score, geometric mean, and Mathews correlation coefficient (MCC):

\begin{equation}
         \text{Accuracy} = \frac{TP + TN}{TP + FP + TN + FN}
    \end{equation}

    \begin{equation}
         \text{Precision} = \frac{TP}{TP + FP}
    \end{equation}

    \begin{equation}
        \text{Recall} = \frac{TP}{TP + FN}
    \end{equation}

    \begin{equation}
         \text{F1-score} = \frac{2 \times \text{Precision} \times \text{Recall}}{\text{Precision} + \text{Recall}}
    \end{equation}

where TP, TN, FN, and FP denote the true positive, true negative, false negative, and false positive values, respectively. The CF-GNN experiments, we evaluated each model using standard performance metrics. The precision, recall, and F1-score metrics vary significantly by class, indicating differences in how well the model predicts each class. This heterogeneity in performance suggests that the model may perform well for some classes and less effectively for others. The geometric mean is computed using a TP rate (TPR) and a TN rate (TNR).

\begin{equation}
    \text{Geometric mean} = \sqrt{\text{TPR} \cdot \text{TNR}} = \sqrt{\frac{\text{TP}}{\text{TP} + \text{FN}} \cdot \frac{\text{TN}}{\text{TN} + \text{FP}}}.
\end{equation}

A higher score for the mentioned criteria indicates better model performance on imbalanced problems. The MCC generalizes the confusion matrix for multiclass classification. The generalized equation is:

\[
\text{\scriptsize{
\(\text{MCC} = \frac{\sum_{k} \sum_{l} \sum_{m} C_{kk} C_{lm} - C_{kl} C_{mk}}{\sqrt{\left( \sum_{k} \left( \sum_{l} C_{kl} \right) \left( \sum_{l \neq k} C_{lk} \right) \right) \cdot \left( \sum_{k} \left( \sum_{l} C_{lk} \right) \left( \sum_{l \neq k} C_{kl} \right) \right)}}\)
}}
\]
where \( k, l, and m \) represent indices corresponding to certain classes and \(C_{kl}\) denotes the elements of the confusion matrix. The MCC is particularly suitable for multiclass imbalanced situations because it ensures a balanced assessment across all classes. The classification mean accuracy (cmA) computes per-class recall and averages it over all $K$ classes.

\begin{equation} \label{cmA}
    \text{cmA} = \frac{1}{K} \sum_{k=1}^{K} \frac{\text{TP}_k}{\text{TP}_k + \text{FN}_k}
\end{equation}

\section{Experiments} \label{4}

This section provides an empirical analysis of the CF-GNN using two 5G network digital twin datasets. This section describes and analyzes the experimental results, including the datasets and evaluation metrics, followed by a detailed evaluation and discussion of the experimental results.

\subsection{Overall performance comparison}

\begin{figure}[htbp]
        \centering
        \includegraphics[width=08cm]{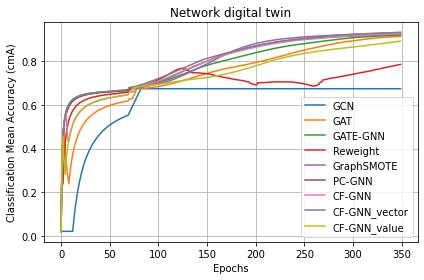}
        \caption{Classification mean accuracy (cmA) comparing network digital twin dataset model performance}
        \label{fig: Perform a}
    \end{figure}

\begin{figure}[htbp]
        \centering
        \includegraphics[width=08cm]{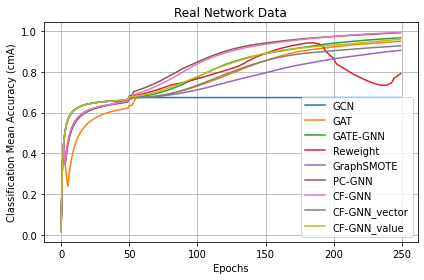}
        \caption{Classification Mean Accuracy (cmA) comparing the real-network model performance}
        \label{fig: Perform c}
    \end{figure}

This section evaluates the effectiveness of the proposed CF-GNN technique, and its performance was compared with state-of-the-art models using the class balanced accuracy metric (cmA) for handling imbalanced classification tasks. Figures \ref{fig: Perform a} and \ref{fig: Perform c} demonstrate that conventional models such as GCN and GAT, struggled with imbalanced datasets, often plateauing at lower accuracy levels. For example, GAT exhibits fluctuations during training, indicating instability, whereas GCN reaches an early performance maximum and fails to adapt effectively to minority classes.

In contrast, other methods, such as PC-GNN and GraphSMOTE, perform better due to enhanced node representations and oversampling approaches. However, other techniques, such as Reweight, have significant instability, particularly in the initial stages of training, suggesting that merely adjusting loss functions is insufficient to address class imbalance effectively. Although GATE-GNN, which aggregates node features using adaptive weights, performs competitively, it is still inconsistent during the training.

\begin{table}[h]
    \centering
    \caption{Classification performance of CF-GNN and GNN-based methods of the two datasets (best results are bold). G-Mean (Abbr. Geometric mean), MCC (Abbr. Mathew correlation coefficient), F1 (Abbr. Macro-F1), Recall (Abbr. Macro-recall) }
    \label{tab:5G core}
    \footnotesize 
    \begin{tabular}{l|l|cccc|cccc}
        \toprule
        \multirow{2}{*}{\rotatebox{90}{\textbf{}}} & \multirow{2}{*}{\textbf{Datasets}} & \multicolumn{4}{c|}{\textbf{Network digital twin datasets}} & \multicolumn{4}{c}{\textbf{Real network datasets}} \\ 
        \cmidrule(lr){3-6} \cmidrule(lr){7-10} 
        & Metrics & G-Mean & MCC & Recall & F1 & G-Mean & MCC & Recall & F1  \\
        \midrule
        \multirow{10}{*}{\rotatebox{90}{\textbf{ Models}}} 
        & GCN & 0.40 & 0.50 & 0.71 & 0.65 & 0.28 & 0.20 & 0.58 & 0.68  \\
        & GAT & 0.62 & 0.79 & 0.70 & 0.76 & 0.22 & 0.52 & 0.64 & 0.75  \\ 
        & PC-GNN & 0.25 & 0.71 & 0.78 & 0.61 & 0.71 & 0.61 & 0.76 & 0.82  \\
        & GraphSMOTE  & 0.23 & 0.71 & 0.78 & 0.71 & 0.45 & 0.88 & 0.87 & 0.88  \\
        & Reweight  & 0.26 & 0.72 & 0.75 & 0.71 & \underline{0.75} & 0.85 & 0.82 & 0.85  \\
        & GATE-GNN  & 0.54 & 0.80 & \textbf{0.87} & \underline{0.82} & 0.63 & 0.95 & 0.94 & 0.95  \\
        \midrule
        & CF-GNN & \textbf{0.75} & \underline {0.83} & 0.70 & 0.73 & \textbf{1.00} & \textbf{1.00} & \textbf{0.98} & \underline{0.96}  \\
        & CF-GNN\textsubscript{eigenvector} & 0.01 & \textbf{0.84} & \underline{0.81} & \textbf{0.89} & 0.00 & 0.59 & 0.43 & 0.42 \\
        & CF-GNN\textsubscript{eigenvalue} & \underline{0.75} & 0.73 & 0.70 & 0.73 & 0.57 & \underline{0.96} & \underline{0.95} & \textbf{0.97} \\
        \bottomrule
    \end{tabular}
\end{table}

We compared the performance of the CF-GNN against all competing GNN-based methods across datasets of network digital twins and real-world networks. Table \ref{tab:5G core} reveals that CF-GNN models outperform all competing GNN-based methods in classification. For instance, considering the F1-score on the real-world network dataset, the CF-GNN outperforms the GCN, GAT, PC-GNN, GraphSMOTE, Reweight, and GATE-GNN by 29\%, 22\%, 6\%, 22\%, 17\%, and 15\% respectively. Similarly, on the network digital twin dataset, the CF-GNN surpasses the GCN, GAT, PC-GNN, GraphSMOTE, Reweight, and GATE-GNN by 8\%, 8\%, 18\%, 6\%, 11\%, and 0\%, respectively. Furthermore, the CF-GNN achieves the highest geometric mean of 0.75, MCC of 0.83, and an overall balanced performance across all metrics, making it the most effective model in both domains. 

To validate the contributions of the CF-GNN spectral components, we conducted an ablation study using two key variants. \textit{CF-GNN\_eigenvalue} applies class-conditioned filtering based solely on graph eigenvalues to emphasize spectral frequency components. This variant captures broad structural variations but lacks localized class-awareness. As observed in the performance plots, it demonstrates stable convergence but slightly underperforms the full CF-GNN on both datasets. In contrast, \textit{CF-GNN\_eigenvector} utilizes eigenvector-based attention, allowing the model to assign importance to feature components aligned with structural class cues. This variant exhibits improved adaptation to minority classes compared to the eigenvalue-only model and performs competitively with the full CF-GNN.

The CF-GNN model achieves superior performance by combining eigenvalue-based spectral filtering with eigenvector-based attention, effectively addressing class imbalance in graph-structured data. Its class-specific filtering mechanism enables the model to extract more discriminative features, particularly for minority classes. This dual spectral approach enhances adaptability and stability, leading to a 2--12\% increase in class-mean accuracy (cmA) over the best baseline models. Overall, CF-GNN strikes an excellent balance between accuracy, generalization, and training stability, making it highly effective for imbalanced graph classification tasks.

\subsection{Imbalanced ratio influence}
The proposed CF-GNN shows strong robustness and adaptability across a wide range of class imbalance scenarios (from 0.1 to 1), as demonstrated in both real network datasets and network digital twin datasets in Tables \ref{4} and \ref{5}. In the experiment, we generate new datasets by randomly sampling different classes according to varying imbalance ratios (IRs). These datasets are then used to evaluate CF-GNN alongside all baseline models.

As shown in Table~\ref{tab:4}, the CF-GNN and its variants show remarkable adaptability across imbalance ratios. At the lowest imbalance ratio of 0.1, the CF-GNN\textsubscript{Eigenvecto} achieves the highest performance (88.47\%), outperforming all other methods, including PC-GNN (83.13\%) and Reweight (80.69\%). This demonstrates that integrating eigenvector aware attention effectively enhances sensitivity to minority-class signals even in highly skewed settings.

When the imbalance ratio increases to 0.3, CF-GNN\textsubscript{eigenvector} continues to lead with 93.38\%, surpassing other strong contenders such as GraphSMOTE (74.36\%) and PC-GNN (86.75\%). The variants CF-GNN\textsubscript{eigenvalue} and CF-GNN\textsubscript{eigenvector} also remain competitive at 92.05\% and 78.71\%, respectively. These results highlight the effectiveness of the proposed class-specific spectral filtering and attention mechanisms.

At an imbalance ratio of 0.5, the CF-GNN\textsubscript{eigenvector} again achieves the highest score (97.36\%), indicating strong mid-level imbalance robustness. The variants CF-GNN and CF-GNN\textsubscript{eigenvector} follow closely (95.38\% and 87.25\%), significantly outperforming traditional methods such as GCN (68.13\%) and GAT (67.25\%), affirming the importance of eigenvector-based feature recalibration in capturing graph semantics.

At 0.7 imbalance ratio, although all models experience some decline, the CF-GNN\textsubscript{eigenvector} and CF-GNN still dominate with 93.87\% and 89.87\%, respectively, while GraphSMOTE (76.65\%) and PC-GNN (84.07\%) show decreased performance. This underscores that CF-GNN variants maintain resilience under increasing imbalance severity.

At an extreme imbalance of 0.9, the performance of CF-GNN peaks at (98.02\%) showcasing its capacity to handle rare classes. The CF-GNN\textsubscript{eigenvector} again exihibits performance at (92.85\%) and GraphSMOTE (93.46\%). The results suggest that while oversampling strategies like GraphSMOTE are helpful in extreme cases, the Fourier-aware learning in CF-GNN still leads overall.

\begin{table}[htbp]
    \centering
    \caption{Real-world network dataset results by imbalance ratios (bold: best, underline: runner-up)}
    \label{tab:4}
    \begin{tabular}{lccccc}
        \toprule
        \textbf{Methods} & \textbf{0.1} & \textbf{0.3} & \textbf{0.5} & \textbf{0.7} & \textbf{0.9} \\
        \midrule
        GCN & 0.6874 & 0.6843 & 0.6813 & 0.6783 & 0.6754 \\
        GAT & 0.6847 & 0.6029 & 0.6725 & 0.7512 & 0.7584 \\
        GraphSMOTE & 0.7126 & 0.7436 & 0.7689 & 0.7665 & \underline{0.9346} \\
        PC-GNN & \underline{0.8313} & 0.8675 & 0.8407 & 0.8453 & 0.8876 \\
        Reweight & 0.8069 & 0.8603 & 0.8879 & 0.7593 & 0.8126 \\
        GATE-GNN & 0.8049 & 0.7891 & 0.7908 & 0.7561 & 0.7707 \\
        CF-GNN & 0.7406 & \underline{0.9205} & \underline{0.9530} & \textbf{0.9887} & \textbf{0.9802} \\
        CF-GNN\textsubscript{eigenvalue}  & 0.7907 & 0.7871 & 0.8725 & 0.7564 & 0.7996 \\
        CF-GNN\textsubscript{eigenvector} & \textbf{0.8847} & \textbf{0.9338} & \textbf{0.9736} & \underline{0.9387} & 0.9285  \\
         
        \bottomrule
    \end{tabular}
\end{table}

\begin{table}[htbp]
    \centering
    \caption{Network digital twin dataset results by imbalance ratios (bold: best, underline: runner-up)}
    \label{tab:5}
    \begin{tabular}{lccccc}
        \toprule
        \textbf{Methods} & \textbf{0.1} & \textbf{0.3} & \textbf{0.5} & \textbf{0.7} & \textbf{0.9} \\
        \midrule
        GCN & 0.4874 & 0.4842 & 0.5825 & 0.67834 & 0.6754 \\
        GAT & 0.5847 & 0.6092 & 0.6724 & 0.7532 & 0.7570 \\
        GraphSMOTE & 0.7126 & 0.7436 & 0.7689 & \underline{0.8665} & \textbf{0.9346} \\
        PC-GNN & \underline{0.8313} & 0.7695 & 0.7908 & 0.7561 & 0.7876 \\
        Reweight & 0.8060 & 0.7177 & 0.7373 & 0.7303 & 0.7763 \\
        GATE-GNN & 0.7549 & 0.7517 & 0.6736 & 0.7510 & 0.7261 \\
        CF-GNN & 0.7406 & \textbf{0.8338} & 0.8538 & \textbf{0.8987} & \underline{0.9085} \\
         CF-GNN\textsubscript{eigenvector} & 0.7847 & \underline{0.8291} & \underline{0.8018} & 0.7561 & 0.7707 \\
         CF-GNN\textsubscript{eigenvalue}  &\textbf{0.8507} & 0.8171 & \textbf{0.9125} & 0.8564 & 0.8996 \\
        \bottomrule
    \end{tabular}
\end{table}

In Table~\ref{tab:5}, the results on the network digital twin dataset further affirm the effectiveness of CF-GNN variants across imbalance conditions.

At an imbalance ratio of 0.1, CF-GNN-Eigenvalue achieves the best accuracy (85.07\%), outperforming CF-GNN-Eigenvector (78.47\%) and the base CF-GNN (74.06\%). PC-GNN (83.13\%) and Reweight (80.60\%) also perform strongly, yet fall short of the spectral variant. This indicates that the eigenvalue-based spectral transformation is especially effective at low imbalance levels on digital twin data.

On the network digital twin dataset, the performance pattern of the CF-GNN remains consistent and notably robust. At imbalance ratio 0.1, the CF-GNN achieves (85.07\%), outperforming CF-GNN\textsubscript{eigenvector} (78.74\%) and the base CF-GNN (74.06\%). The PC-GNN (83.13\%) and Reweight (80.60\%) also perform strongly, yet fall short of the spectral variant. This indicates that the eigenvalue-based spectral transformation is especially effective at low imbalance levels on digital twin data.

At an imbalance ratio of 0.3, the CF-GNN outperforms baseline models with 83.38\%, indicating its solid performance under moderate imbalance. The CF-GNN\textsubscript{eigenvector} also performs competitively (82.91\%), reinforcing the utility of eigenvector-based attention in moderately skewed distributions and confirming that attention on eigenvectors enables better feature localization on simulated graphs structured data.

As imbalance increase to 0.5, the CF-GNN\textsubscript{eigenvalue} dominates performance with (91.25\%), whereas CF-GNN\textsubscript{eigenvector} and the base CF-GNN follow closely (80.18\% and 85.38\%, respectively). This further validates the spectral filtering capability of CF-GNN over reweighting and synthetic sampling methods for nodes classification task.

At 0.7 imbalance, the CF-GNN\textsubscript{eigenvalue} retains the lead (85.64\%) over CF-GNN (80.85\%) and CF-GNN\textsubscript{eigenvector} (75.61\%), outperforming GCN (67.83\%), GAT (75.32\%), and GraphSMOTE (86.65\%). These results demonstrate how eigenvalue-guided transformations offer consistent advantages in highly imbalanced settings.

At the extreme 0.9 imbalance, CF-GNN performs best at 90.85\%, followed closely by CF-GNN\textsubscript{eigenvalue} (89.96\%) and CF-GNN\textsubscript{eigenvector} (77.07\%) highlighting that graph spectral filtering augmented with eigenvector attention offers robust generalization even under extreme class imbalance. In particular, GraphSMOTE decreases in performance (67. 53\%), suggesting that synthetic oversampling is less effective in digital twin datasets compared to real-world network data.

\subsection{Multi-class classification results}

\begin{figure}[htbp]
    \centering
    \includegraphics[width=10cm]{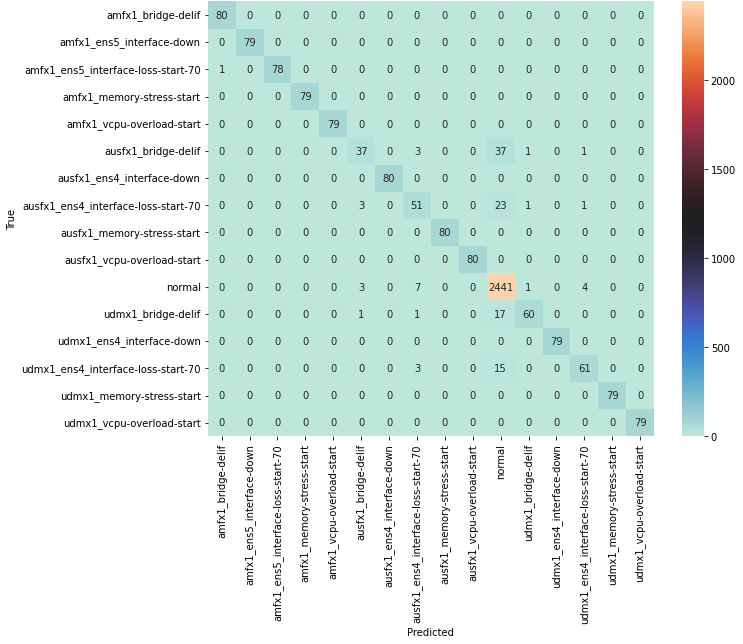}
    \caption{Confusion matrix of the network of digital twin environment (Domain A)}
    \label{fig: Network of Digital Twin (Domain A)}
\end{figure}

The classification reports provide the precision, recall, and F1-score for every class in the datasets in Domain A and Domain C comprehensively analyzing performance across domains. Table \ref{tab A} presents slightly lower performance on the training data from the network digital twin environment in Domain A is slightly lower compared to the real network environments in Domain C, with an average F1-score of 0.93. Precision and recall values are also marginally lower, around 0.94, indicating that the model performs well but may not identify some classes accurately. Precision measures the proportion of TP predictions out of all positive predictions (TPs and FPs) made by the model. In the Domain A dataset, Classes 1, 3, 6, 8, 9, 12, and 14 achieved perfect precision (1.00), indicating that the model correctly identified all instances belonging to these classes without any false positives. However, Classes 5, 7, and 13 have lower precision values, suggesting that the model predictions for these classes include increased FPs.
In the Domain C dataset (Figure \ref{fig: Network of Digital Twin (Domain A)}), most classes achieved high precision values, with Classes 1-4, 6-10, and 12-15 achieving perfect precision (1.00). However, Class 0 has a lower precision value, indicating that the model predictions include increased FPs.

\begin{table*}[htbp]
    \centering
    \caption{Classification results for training in a network digital twin Environment (Domain A) and real-world network data (Domain C)}
    \label{tab A}
    \begin{tabular}{c|c|ccc|ccc}
            \hline \hline
            & \multicolumn{1}{c|}{Datasets} & \multicolumn{3}{c}{Domain A} & \multicolumn{3}{c}{Domain C} \\  \cline{1-8}
              Classes & Sub-Types & Pre  & Recall & F1 & Pre & Recall & F1 \\
             \hline
0 & amfx1\_bridge-delif & 0.99 & 0.86 & 0.92 & 0.88 & 0.70 & 0.78 \\
1 & amfx1\_ens5\_inter-down & 1.00 & 1.00 & 1.00 & 1.00 & 1.00 & 1.00 \\
2 & amfx1\_ens5\_inter-loss-70 & 0.91 & 0.87 & 0.89 & 1.00 & 0.90 & 0.95 \\
3 & amfx1\_memory-stress-start & 1.00 & 0.99 & 0.99 & 1.00 & 1.00 & 1.00 \\

4 & amfx1\_vcpu-overload-start & 0.95 & 0.91 & 0.93 & 1.00 & 0.90 & 0.95 \\
5 & ausfx1\_bridge-delif & 0.68 & 0.52 & 0.59 & 0.82 & 0.90 & 0.86 \\
6 & ausfx1\_ens5\_inter-down & 1.00 & 1.00 &  1.00 & 1.00 & 1.00 & 1.00 \\
7 & ausfx1\_ens5\_inter-loss-70 & 0.69 & 0.25 & 0.37 & 1.00 & 1.00 & 1.00 \\
8 & ausfx1\_memory-stress-start & 1.00 & 1.00 & 1.00 & 1.00 & 1.00 & 1.00 \\
9 & ausfx1\_vcpu-overload-start & 1.00 & 1.00 & 1.00 & 1.00 & 1.00 & 1.00 \\
10 & normal & 0.94 & 1.00 & 0.97 & 0.97 & 1.00 & 0.98 \\
11 & udmx1\_bridge-delif & 0.92 & 0.58 & 0.71 & 0.83 & 0.50 & 0.62 \\
12 & udmx1\_ens5\_inter-down & 1.00 & 1.00 & 1.00 & 1.00 & 0.90 & 0.95 \\
13 & udmx1\_ens5\_inter-loss-70 & 0.95 & 0.25 & 0.40 & 1.00 & 0.80 & 0.89 \\
14 & udmx1\_memory-stress-start & 1.00 & 1.00 & 1.00 & 1.00 & 1.00 & 1.00 \\
15 & udmx1\_vcpu-overload-start & 0.92 & 1.00 & 0.96 & 1.00 & 1.00 & 1.00 \\
             \hline
      \hline  
        & \textbf{Macro avg}  & 0.99 & 0.83 & 0.86 & 0.97 & 0.91 & 0.94 \\
      &  \textbf{weighted avg}  & 0.99 & 0.96 & 0.97 & 0.97 & 0.97 & 0.97 \\
    \hline
    \end{tabular}
\end{table*}


\begin{figure}[htbp]
    \centering
    \includegraphics[width=10cm]{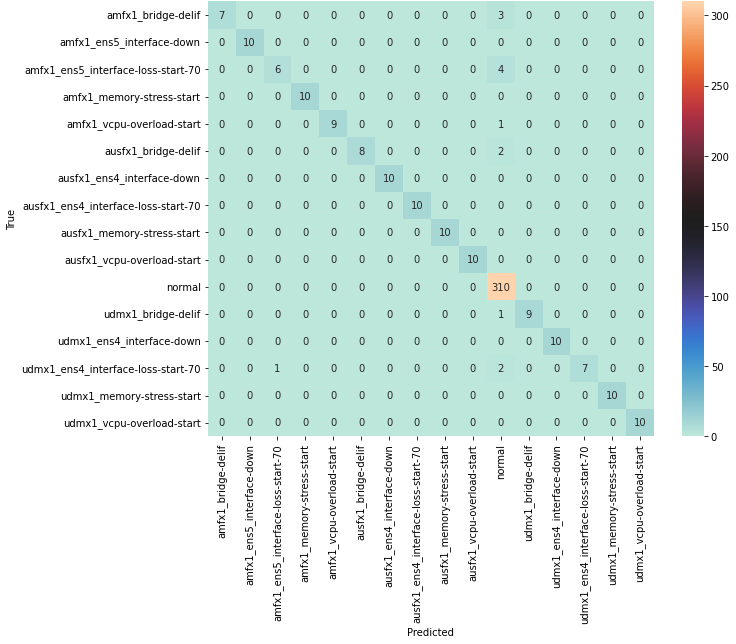}
    \caption{Confusion matrix obtained from the real network (Domain C)}
    \label{fig: Real Network Data (Domain C)}
\end{figure}

Recall, also known as sensitivity, measures the proportion of TP predictions out of all actual positive instances (TPs and FNs) in the dataset. In Figure \ref{fig: Real Network Data (Domain C)}, classes 1, 3, 6, 8, 9, 12, 14, and 15 achieved perfect recall (1.00), indicating that the model correctly identified all instances of these classes. However, Classes 0, 2, 4, 5, 7, 11, and 13 have lower recall values, suggesting that the model missed some instances of these classes. The F1-score, the harmonic mean of the precision and recall, provides a balanced measure of the two metrics, which is helpful for datasets with class imbalance. The weighted average F1-score for the Domain A dataset is 0.93, indicating good overall performance across all classes. The weighted-average F1-score for the Domain C dataset is 0.97, reflecting excellent performance across all classes. 

\begin{figure}[htbp]
        \centering
       \includegraphics[width=11cm]{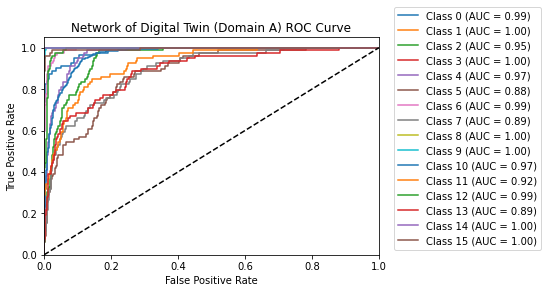}
        \caption{Receiver operating characteristic (ROC) curve analysis of the network of digital twins (Domain A)}
        \label{fig: ROC Curve A}
    \end{figure}
    
    \begin{figure}[htbp]
        \centering
        \includegraphics[width=11cm]{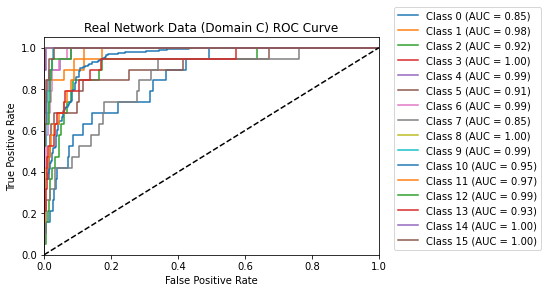}
        \caption{Receiver operating characteristic (ROC) curve analysis of the real network (Domain C)}
        \label{fig: ROC Curve C}
    \end{figure}
 
Overall, Figure \ref{fig: ROC Curve C} and Figure \ref{fig: ROC Curve A} demonstrate robust performance in terms of precision, recall, and F1-score results for most classes, indicating that the models effectively classify instances across categories. The results suggest that the model accurately identifies and classifies most data points. However, Class 0 has the lowest recall (0.63) compared to the others, meaning the model might be missing a sizable portion of actual Class 0 instances, misclassifying them as other categories. Similarly, Class 5 has a lower recall (0.68) and F1 score (0.81), and Class 13 has a slightly lower recall (0.95) than others. Further investigation into Classes 0, 5, and 13 is recommended to improve the classification of these classes and enhance the overall model effectiveness.

\section{Conclusion and Future work} \label{5}
This paper proposes the CF-GNN model to address class imbalances in node classification, particularly in 5G network digital twin datasets. The proposed technique integrates a spectral filtering method for multiclass classification, ensuring a more balanced learning process. The critical analysis of the literature identified critical challenges related to dataset imbalances, where majority-class dominance causes suboptimal embeddings for minority-class nodes. This imbalance complicates classification, particularly as the number of classes increases. 
We examined how loss function consistency varies across different subgraphs, demonstrating that incorporating class frequency and the node degree into the loss formulation improves the distribution balance while preserving the graph structure. Furthermore, we analyzed how nodes contribute to learning class-specific filter parameters. The findings suggest that some nodes play a more significant role in updating filter parameters, influencing the ability of the model to differentiate between classes effectively.

Experimental results on a real network and digital twin datasets validated the effectiveness of the CF-GNN in mitigating class imbalance and enhancing minority-class classification. Future work should further investigate the role of subclass structures within minority groups and refine GNN designs that employ data-level techniques to enhance the classification performance on imbalanced graph data.

\section{Declaration of competing interest}
The authors declare that they have no known competing financial interests or personal relationships that could have appeared
to influence the work reported in this article.

\section{Acknowledgement}
This work was supported by the Institute of Information and Communications Technology Planning and Evaluation (IITP) grant, funded by the Korean government (MSIT), under grant number (RS-2024-00345030), Development of Digital Twin-Based Network Failure Prevention and Operation Management Automation Technology.

\section{Data availability}

Data are available at https://aiforgood.itu.int/event/2023-japan-challenge-network-failure-classification-using-network-digital-twin.


\bibliographystyle{elsarticle-num}
\bibliography{main}
\end{document}